\documentclass[10pt,twocolumn,letterpaper]{article}

\usepackage[pagenumbers]{cvpr} 

\usepackage{graphicx}
\usepackage{amsmath}
\usepackage{amssymb}
\usepackage{booktabs}
\usepackage{xcolor}
\usepackage{algorithm}
\usepackage{algpseudocode}
\usepackage{soul}
\usepackage{booktabs}
\pdfoutput=1

\usepackage[pagebackref,breaklinks,colorlinks]{hyperref}
\usepackage{array}
\usepackage{multirow}

\usepackage[capitalize]{cleveref}
\crefname{section}{Sec.}{Secs.}
\Crefname{section}{Section}{Sections}
\Crefname{table}{Table}{Tables}
\crefname{table}{Tab.}{Tabs.}

\def\cvprPaperID{***} 
\def\confName{CVPR}
\def\confYear{2023}

\begin{document}

\newcommand{\todo}[1]{\textcolor{red}{#1}}
\newcommand{\loss}{\mathcal{L}}
\newcommand{\T}{\mathcal{T}}
\newcommand{\g}{\mathbf{g}}
\newcommand{\W}{\mathbf{W}}
\newcommand{\C}{\mathcal{C}}
\newcommand{\AC}{\mathcal{A}\mathcal{C}}
\newcommand\norm[1]{\left\lVert#1\right\rVert}
\newcolumntype{P}[1]{>{\centering\arraybackslash}p{#1}}

\newcommand{\Lijun}[1]{\textcolor{blue}{{\it [Lijun: #1]}}}
\newcommand{\hui}[1]{\textcolor{blue}{{\it [Hui: #1]}}}

\title{Structured Pruning for Multi-Task Deep Neural Networks }

\author{Siddhant Garg\\
University of Massachusetts Amherst\\
{\tt\small siddhantgarg@umass.edu}
\and
Lijun Zhang\\
University of Massachusetts Amherst\\
{\tt\small lijunzhang@cs.umass.edu}
\and
Hui Guan\\
University of Massachusetts Amherst\\
{\tt\small huiguan@cs.umass.edu}
}
\maketitle

\begin{abstract}

    Although multi-task deep neural network (DNN) models have computation and storage benefits over individual single-task DNN models, they can be further optimized via model compression.
   Numerous structured pruning methods are already developed that can readily achieve speedups in single-task models, but the pruning of multi-task networks has not yet been extensively studied.
   In this work, we investigate the effectiveness of structured pruning on multi-task models. 
   We use an existing single-task filter pruning criterion and also introduce an MTL-based filter pruning criterion for estimating the filter importance scores. 
   We prune the model using an iterative pruning strategy with both pruning methods. 
   We show that, with careful hyper-parameter tuning, architectures obtained from different pruning methods do not have significant differences in their performances across tasks when the number of parameters is similar. 
    We also show that iterative structure pruning may not be the best way to achieve a well-performing pruned model because, at extreme pruning levels, there is a high drop in performance across all tasks. 
    But when the same models are randomly initialized and re-trained, they show better results.  
\end{abstract}

\section{Introduction}
\label{sec:intro} 



Multi-task learning (MTL) addresses multiple machine learning tasks simultaneously by creating a single multi-task deep neural network (DNN) \cite{sener2018multi-hard-2}.
Due to parameter sharing, a multi-task DNN model is more computation and memory efficient compared to multiple single-task models. 
The MTL framework is extremely useful for resource-constrained devices like smartphones, wearables, and self-driving cars that host AI-powered applications but have low memory resources and strict latency requirements. 
For example, in self-driving cars, the model needs to recognize traffic lights, objects, and lanes based on the input signal \cite{sd-mtl-1-phillips2021deep}. 



On the other hand, network pruning \cite{optimal=brain-damage, lth-1, pruning-filters} is a long-standing and effective method to compress DNNs. 
It aims to detect the importance of model parameters and remove the ones that tend to have the least significance on the performance. 
Approaches for network pruning can be classified as unstructured pruning methods \cite{lth-1, optimal=brain-damage} that mask individual weights in the network, and structured pruning methods \cite{pruning-filters, taylor-pruning} that remove complete filters and directly lead to efficient deep neural models without requiring specialized hardware for sparse structures. 
Network pruning is studied rigorously in literature for single-task models and there exist many pruning criteria based on weight magnitude \cite{lth-1, pruning-filters}, connection sensitivity \cite{lee2018snip, rigl-evci2020rigging}, and even learning-based pruning methods \cite{ding2021resrep-learn-1, hou2022chex-learn-2}. 

However, the study on the effectiveness of structured pruning methods on multi-task models is sparse. 
A few works like \cite{cheng2021multi-filter-index-share, he2021pruning-pam} propose weight sharing and merging strategies for constructing a multi-task model from multiple single-task models such that there is a minimum conflict between tasks. 
Then the constructed multi-task model could be effectively pruned with single-task pruning methods. 
But these works show very similar results when comparing their proposed pipeline with the baseline single-task pruning methods. 
Another work \cite{sun2022disparse}, proposed a method to directly prune MTL networks but while pruning those models with single-task pruning baselines, they did not try different hyperparameter settings to retrain/fine-tune the pruned models. 
It could be possible that different hyperparameter settings would work better because different pruning methods lead to different architectures. 

In this work, we investigate the effectiveness of structured pruning on multi-task DNNs. We apply two structured pruning methods to prune multi-task models and show that regardless of the method used, we can obtain similar results from the pruned models with the same number of parameters. 
Specifically, we use an existing single-task pruning method as well as introduce another MTL-based pruning criterion. 
The proposed criterion is called CosPrune and it identifies and prunes the convolutional filters that have conflicts between tasks. 
It uses pairwise cosine similarity between the task-specific gradients that flow through the filter during back-propagation. 
We accumulate this similarity score for some training iterations for every filter in the multi-task model. 
The filters with the least accumulated scores are pruned away. 
In contrast, the single-task pruning method is called Taylor Pruning \cite{taylor-pruning} which is a popular gradient-based pruning method. 
It determines the importance of the filter by looking at the increase in the loss function if that filter is removed. 
The Taylor pruning importance score is also accumulated over some training iterations before pruning.

We start our analyses by using the iterative pruning and fine-tuning strategy which repeatedly prunes a small proportion of filters and fine-tunes the multi-task model to gain back the lost performance \cite{lth-1, liu2018rethinking}. 
Using this strategy we get consistently better results with CosPrune against Taylor pruning across all the tasks. The multi-task model achieved higher GFLOPs/parameter reduction with CosPrune without performance loss across all the tasks. This shows that the proposed CosPrune criterion coupled with iterative pruning is a reasonable method for pruning multi-task models. 


\textbf{However, when we re-train the pruned models independently with random initialization, we observe that they can give relatively better results on all the tasks when compared to the corresponding fine-tuning stage of iterative pruning.} The key is to determine good learning rates for each of the pruned models. 
Using the same learning rates is not the best strategy to compare different pruned architectures. This is because every model has different layer-wise configurations and so, an optimal hyper-parameter setting for one model may not be best for the other models with different architectural designs. This type of analysis is generally not done in the existing literature -- the same training settings are used for the dense model as well as the pruned models. A study on single-task structured pruning \cite{liu2018rethinking} also shows that re-training the pruned models from random initialization can lead to better results than fine-tuning the pruned architectures in the iterative pruning setting. 

Furthermore, the pruned architectures from different pruning methods give similar results to each other at the same parameter level after re-training. 
\textbf{There are no consistent winners with respect to different pruning criteria, which is contrary to what we observe in the case of iterative pruning. }
There are also some recent works \cite{random-li2022revisiting, random-liu2022unreasonable} in the context of single-task neural networks where random channel pruning is able to match the results of the dense model under appropriate settings. Similarly, another work \cite{liu2018rethinking} used different pruning methods on various architectures to show that randomly re-initializing the pruned models can match the performance of the respective unpruned models but they did not compare those different pruning strategies on the same model. 

We go beyond existing works and apply different structured pruning methods to the same multi-task model. \textbf{We emphasize that the pruned MTL model obtained from any reasonable pruning method can perform well if it is trained from random initialization with its optimal learning rate.} 
And we would like researchers to pay attention to the re-training comparisons with sufficient hyper-parameter tuning when they try to propose a new pruning method.

\section{Related Works}

\textbf{Multi-Task Learning}: Deep MTL networks afford numerous benefits in terms of computation (storage and latency) \cite{kim2021mila-latency}, knowledge sharing between tasks \cite{abdollahzadeh2021revisit-knowledge}  and improving generalization in the learned representations \cite{misra2016cross-general} due to which they have applications across many domains like Computer Vision \cite{nyudepthnormal, hu2021unit-cv1, guo2018dynamic-cv2, ghiasi2021multi-cv3}, Robotics \cite{williams2008multi-robo-1, yu2020meta-robo-2}, and Reinforcement Learning \cite{rl-mtl-1-teh2017distral, rl-mtl-2-sodhani2021multi, wilson2007multi-rl-3, yang2020multi-rl-4}. The most common MTL framework is hard parameter sharing \cite{caruana1993multitask-hard-1, sener2018multi-hard-2} where a backbone network is shared among all tasks and with individual task-specific heads. Soft parameter sharing, on the other hand, has different sets of parameters for individual tasks \cite{misra2016cross-general, sun2020adashare} and various methods are used to effectively combine information from them \cite{lu2017fully-c1, kendall2018multi-c2, ruder122017learning-c3} to make predictions.  

\textbf{Deep Neural Network Pruning}: It has long been established that the deep neural networks are highly over-parameterized \cite{optimal=brain-damage, brain-surgeon-second-derivative,pruning-filters, bayesian-compression, weights-and-connections}, and more than $90\%$ of the connections can be pruned away without losing accuracy. Unstructured pruning methods for single-task models \cite{lth-1, wt-rewind} can achieve high sparsity with latency improvements with accelerated hardware for sparse neural networks \cite{sparse-hardware-zhu2019sparse}. On the other hand, structured pruning can directly lead to computational benefits by removing whole filters in the case of convolutional layers \cite{pruning-filters}. Gradient-based metrics for estimating the importance scores of the filters have recently become popular over magnitude-based criteria \cite{pruning-filters}. For example, Taylor Pruning \cite{taylor-pruning} uses the dot product between the filter weights with its gradient to approximate the change in loss function that could occur if that filter is masked. Another example is SNIP \cite{lee2018snip}, which determines the connection sensitivity using a similar importance metric as Taylor Pruning but it is done only once at initialization to apply single-shot pruning. Furthermore, random channel pruning methods \cite{random-li2022revisiting, random-liu2022unreasonable} are also shown to perform well under appropriate training conditions. 

\textbf{Pruning Multi-task Neural Networks}: There is very limited study on pruning MTL neural networks but nonetheless it is an important problem because of its tremendous potential in deriving efficient deep learning models. At the same time, it is also a difficult problem because of the coexistence of complex task relationships in the shared parameter space and redundancy in the neural network. A recent work on MTL pruning is DiSparse \cite{sun2022disparse} which aims to disentangle task relationships to find the unanimously least important filters across all the tasks and prune them. Another work called PAM \cite{he2021pruning-pam} propose a method to merge single-task models into an MTL network such that the resulting MTL network can be safely pruned while considering the tasks' relatedness. Similarly, other works like \cite{cheng2021multi-merge-1, he2018multi-merge-2} propose different merging strategies for the construction of computationally efficient MTL networks. All of these existing  works do baseline comparisons by using single-task pruning methods to prune their MTL networks. However, they use only the hyperparameters that are specified in the original study on single-task pruning methods irrespective of the different model architectures and sparsity ratios. But in our work, we do extensive analyses and hyperparameter tuning for both the pruning methods involved to show their true effectiveness in pruning deep MTL networks. 


\section{Pruning Methods}

In this section, we will define our Multi-Task Learning model and the proposed CosPrune importance score estimation method as well as review the Taylor Pruning importance score. 

\textbf{Multi-task framework}: Given a set of $T$ tasks $\T = \{\tau_1, \dots, \tau_T\}$, an MTL dataset with inputs, $\mathcal{X}$, and the corresponding labels $\mathcal{Y} = \{Y_1, \dots, Y_T\}$ where $Y_t$ is set of labels for task $\tau_t \in \T$, we want to learn the mapping $f_{\Theta}: \mathcal{X} \rightarrow \mathcal{Y}$, where $\Theta = \theta_s \cup \{\theta_{t}\}_{t=1}^{T}$. Here $\theta_s$ is the set of model parameters that are shared by all the tasks and $\theta_t$ is the set of parameters only for task $\tau_t$. From now, we will denote $\{\theta_{t}\}$ as the set of all the task-specific parameters for all the tasks collectively and omit values of $t$ for abbreviation. The parameters $\Theta$ are trained by minimizing the multi-task loss function given by 
\begin{align}
    \mathcal{L}(\mathcal{X}, \mathcal{Y}, \Theta) &= \sum_{t=1}^T\ell_t \left(\mathcal{X},\mathcal{Y}, \theta_s, \theta_t \right)
\end{align}
where $\ell_t \left(\mathcal{X},\mathcal{Y}, \theta_s, \theta_t \right)$ is the loss function for the task $\tau_t$. 

\textbf{CosPrune Importance Score Estimation}: Let $\W^{k\times k\times c} \in \Theta$ be a convolutional filter with kernel size $k$, and $c$ channels, then the weights of this filter will be optimized using the gradient descent given by equation \ref{eq:grad-descent},
\begin{align}
\label{eq:grad-descent}
    \W \leftarrow \W - \eta \nabla_{\W} \mathcal{L}(\mathcal{X}, \mathcal{Y}, \Theta)
\end{align}
where $\eta$ is the learning rate and $\nabla_{\W} \mathcal{L}(\mathcal{X}, \mathcal{Y}, \Theta)$ is the gradient of the total loss function with respect to $\W$. Let $\g_{\T}^{\W} = \nabla_{\W} \mathcal{L}(\mathcal{X}, \mathcal{Y}, \Theta)$, then
\begin{align}
\label{eq:total-gradient}
    \g^{\W}_{\T} &= \sum_{t=1}^T \g^{\W}_{t}
\end{align}
where $\g^{\W}_{t}= \nabla_{\W} \ell_t \left(\mathcal{X},\mathcal{Y}, \theta_s, \theta_t \right)$. $\g^{\W}_{\T}$ and $\g^{\W}_{t}$ can be denoted as the total gradient and task-$\tau_t$ gradient respectively with respect to the parameter $\W$. Note that the total gradient is the vector sum of all the individual task gradients. These individual gradients can have high differences in magnitudes as well as they can point to different directions with negative cosine similarity between them. This can lead to conflicts between different tasks and that could be detrimental to the optimization process \cite{pcgrad-yu2020gradient}. There have been several works like PCGrad \cite{pcgrad-yu2020gradient}, MGDA \cite{opt-1-desideri2012multiple}, CAGrad \cite{opt-2-liu2021conflict} that apply various methods to remove the conflicts in the optimization process. In our work, we prune the filters that have the highest degree of accumulated conflicts but optimize our MTL model with the total gradient.  This helps us to achieve the highest degree of optimization in terms of computation and performance. 

\begin{algorithm}[t]
\caption{MTLCosPrune}\label{alg:mtl-cos-prune}
\begin{algorithmic}
\State \textbf{Input}: Dataset: $\mathcal{X}$, $\mathcal{Y}$ with $T$ tasks; Model Parameters: $\W \in \Theta$; Fine-tune Iterations -- $I$; $\#$Filters to prune at each pruning step -- $P$, 
\State \textbf{Initialize} $\Theta$, 
\While{$\#$Unpruned Filters $> threshold$}

    Set $\AC(\W) \leftarrow 0$ $\forall \W \in \Theta$ (accumulated scores)

    \For{$i = 1, 2, \dots, I$}
    
        $\mathbf{x}_i \leftarrow $Batch Inputs
    
        $\mathbf{y}_i \leftarrow $Batch Labels


        $\g^{\W}_{t}= \nabla_{\W} \ell_t \left(\mathbf{x}_i,\mathbf{y}_i, \theta_s^i, \theta_t^i \right)$ $\forall t$, $\forall \W \in \theta_s^i$
        
        Get $\C(\W, \{\g^{\W}_{t}\})$ using \ref{eq:cos-prune} $\forall t$, $\forall \W \in \theta_s^i$

        Update $\AC(\W) \leftarrow \AC(\W) + \C(\W, \{\g^{\W}_{t}\})$ $\forall t$, $\forall \W \in \theta_s^i$

        Compute $\g^{\W}_{\T} = \sum_{t=1}^T \g^{\W}_{t}$

        Update $\W \leftarrow \W - \eta \g^{\W}_{\T}$, $\forall t$, $\forall \W \in \theta_s^i$
        
    \EndFor
    
    Prune $P$ filters $\W$ with lowest $\AC(\W)$ scores
    
\EndWhile
    \end{algorithmic}
\end{algorithm}

To calculate the degree of conflict, we define the CosPrune Importance score, as the sum of pairwise cosine similarity between all the tasks for the parameter $\W$. Let $\Tilde{\g}^{\W}_{t} = \frac{\g^{\W}_{t}}{\norm{\g^{\W}_{t}}}$ be the normalized task gradient, and $\{\g^{\W}_{t}\}$ be the set of all the task gradients with respect to $\W$, then $\C\left(\W, \{\g^{\W}_{t}\} \right)$ can be calculated using:
\begin{align}
\label{eq:cos-prune}
    \C\left(\W, \{\g^{\W}_{t}\}\right) &= \sum_{(\tau_i, \tau_j) \in \T\times \T} \Tilde{\g}^{\W}_{i}. \Tilde{\g}^{\W}_{j}
\end{align}
We only prune the shared parameters, $\W\ \in \theta_s$, therefore $\g^{\W}_{t} \neq \mathbf{0}$ for all $t=1, \dots, T$.  The value of $\C(.,.)$ will be higher when the pairwise gradients are in agreement and it will lower if tasks are in conflict. For calculating the final importance score of a filter, we keep on adding the CosPrune scores for all the filters for a fixed number of training iterations. After that, we rank the filters according to their accumulated importance scores and prune the lowest-scoring filters. After pruning, we reset the accumulated scores to zero and fine-tune the model again along with collecting the CosPrune scores of the remaining unpruned filters. The details are provided in Algorithm \ref{alg:mtl-cos-prune}.

\textbf{Taylor Pruning \cite{taylor-pruning}:} It is a structured pruning method for single-task convolution neural networks which serves as a popular baseline for modern gradient-based pruning methods. In this method, the importance score of a filter is defined by the squared change in the loss function induced by removing the filter. To make the computation efficient, the squared change is approximated by the first-order Taylor expansion which simplifies to a dot product between the filter weights and its gradient. Let the importance score for the filter $\W$ be $\mathbf{I}$, then it is given by
\begin{align}
    \mathbf{I} = \W . \g^{W}_{\T}
\end{align}
where $\g^{\W}_{\T}$ is the total gradient passing through filter $\W$ during backpropagation as define in equation \ref{eq:total-gradient}. 

We experiment with both CosPrune and Taylor pruning methods and compare their performance in effectively pruning the MTL models.

\section{Experiments and Results}


In this section, we provide the details of all the experiments and provide quantitative evidence of our claims. 

\subsection{Setup}

\textbf{Model Architecture}: For our multi-task model, we use a hard parameter-sharing paradigm where the backbone parameters are shared across all the tasks and individual classification heads are used for each task. MTL backbone comprises of VGG-16 \cite{vgg} model without the last fully-connected layers and MTL classification heads use Artrous Spatial Pyramid Pooling (ASPP) heads \cite{aspp} for each task. The backbone contains approximately $71M$ parameters and each ASPP head has approximately $13.4M$ parameters. In all of our experiments, we only prune the backbone but all the results are reported with the total model parameters (backbone + all task heads) which are approximately $84.12M$ parameters.  

\textbf{Multi-Task Learning (MTL) Dataset}: In this work, NYUv2 \cite{nyusegment} dataset  is used for all of the experiments. It is a popular Multi-Task Learning MTL dataset with densely labeled images recorded from RGB and Depth cameras of Microsoft Kinect. It contains three tasks -- Semantic Segmentation, Depth Estimation, and Surface Normal Prediction whose ground truth labels are defined in \cite{nyudepthnormal}. 

\textbf{Evaluation Metrics}: Semantic segmentation is evaluated using mean Intersection over Union (mIoU) and the Pixel accuracy (both the higher the better). Depth Estimation is evaluated using absolute and relative errors calculated using the L1 loss between the ground truth and predictions (both the lower the better). For this task, we also report the relative difference between the prediction and ground truth via the percentage of $\delta = \max \left(\frac{y_{pred}}{y_{gt}}, \frac{y_{gt}}{y_{pred}}\right)$ within threshold $1.25, 1.25^2$, and $1.25^3$ \cite{depth-threshold} (the higher the better). Surface Normal Prediction is evaluated using the Mean and Median Angle errors calculated by cosine similarity loss (the lower the better). We also report the percentage of pixels whose predictions are within $11.25^{\circ}$, $22.5^{\circ}$, and $30^{\circ}$ to the ground truth (higher the better)\cite{nyudepthnormal}. Due to space constraints and a high number of pruned model configurations, we report Pixel Accuracy for Semantic Segmentation, relative error for Depth Estimation, and Angle Mean for Surface Normal Prediction as the main evaluation metrics, for both pruning methods. All other metrics are reported in the supplementary material 

\textbf{Loss Functions and Model Training}: For training on the NYUv2 dataset, the model minimizes the sum of losses from the individual tasks with equal weightage. Semantic segmentation uses cross-entropy loss, Surface Normal Prediction uses cosine similarity loss and Depth Estimation uses L1 loss as its training signals. We start with a randomly initialized MTL model with VGG-16 backbone and ASPP heads and train the model for 500 epochs. We use a batch size of $16$, and $0.0001$ learning rate with cosine scheduling \cite{cosineschedule}. We use AdamW \cite{adamw} optimizer for the iterative pruning experiments. 

\begin{table*}[htb]
  \centering
  \tabcolsep=0.05cm
  \begin{tabular}{P{0.05\linewidth}P{0.05\linewidth}|P{0.075\linewidth}P{0.075\linewidth}P{0.075\linewidth}|P{0.075\linewidth}P{0.075\linewidth}P{0.075\linewidth}|P{0.075\linewidth}P{0.075\linewidth}P{0.075\linewidth}}
    \toprule
    \multicolumn{2}{P{0.1\linewidth}|}{$\#$Params} & \multicolumn{3}{|p{0.25\linewidth}|}{\centering Semantic Segmentation \\ Pixel Accuracy $(\%)$ $\uparrow$} & \multicolumn{3}{|p{0.25\linewidth}|}{\centering Depth Estimation \\ Relative Error $\downarrow$} & \multicolumn{3}{|p{0.25\linewidth}}{\centering Surface Normal \\ Angle Mean $\downarrow$} \\
    \midrule
     $\#(M)$ & $\%_R$ & Taylor & CosPrune & $\%\Delta\uparrow$ & Taylor & CosPrune & $\%\Delta\downarrow$ & Taylor & CosPrune & $\%\Delta (\downarrow)$ \\
    \midrule
    $59.9$ & $28.7$ & $\mathbf{45.96}$ & $45.91$ & $-\phantom{1}0.11$ & $0.2616$ & $\mathbf{0.2614}$ & $\mathbf{-\phantom{1}0.08}$ & $18.54$ & $\mathbf{18.37}$ & $\mathbf{-0.92}$ \\
    $42.7$ & $49.2$ & $45.66$ & $\mathbf{45.70}$ & $\mathbf{+\phantom{1}0.09}$ & $\mathbf{0.2666}$ & $0.2670$ & $+\phantom{1}0.17$ & $18.35$ & $\mathbf{18.22}$ & $\mathbf{-0.73}$ \\
    $21.2$ & $74.8$ & $43.02$ & $\mathbf{44.86}$ & $\mathbf{+\phantom{1}4.27}$ & $0.2862$ & $\mathbf{0.2777}$ & $\mathbf{-\phantom{1}2.98}$ & $18.19$ & $\mathbf{17.19}$ & $\mathbf{-1.53}$ \\
    $17.6$ & $79.1$ & $41.56$ & $\mathbf{42.98}$ & $\mathbf{+\phantom{1}3.41}$ & $0.3384$ & $\mathbf{0.2829}$ & $\mathbf{-16.42}$ & $18.16$ & $\mathbf{18.03}$ & $\mathbf{-0.71}$ \\ 
    $14.3$ & $83.0$ & $35.83$ & $\mathbf{41.70}$ & $\mathbf{+16.39}$ & $0.3896$ & $\mathbf{0.3276}$ & $\mathbf{-15.93}$ & $18.65$ & $\mathbf{18.00}$ & $\mathbf{-3.45}$ \\
    $13.4$ & $84.0$ & $34.88$ & $\mathbf{36.08}$ & $\mathbf{+\phantom{1}3.43}$ & $0.3928$ & $\mathbf{0.3747}$ & $\mathbf{-\phantom{1}4.59}$ & $18.82$ & $\mathbf{18.39}$ & $\mathbf{-2.26}$ \\
    \bottomrule
  \end{tabular}
  \caption{Iterative pruning results at selected iterations. $\# (M)$: \emph{Number of model parameters in millions}. $\%_R$: \emph{$\%$ parameter reduction with respect to full model}. $\%\Delta \uparrow (\downarrow)$: \emph{$\%$ increase (decrease) with respect to Taylor pruning}.  $\uparrow$: \emph{Higher the better} $\downarrow$: \emph{Lower the better}. }
  \label{tab:random-iter-init-all}
\end{table*}

\begin{figure*}[htb]
  \centering
  \begin{subfigure}{0.33\textwidth}
    \includegraphics[scale=0.35]{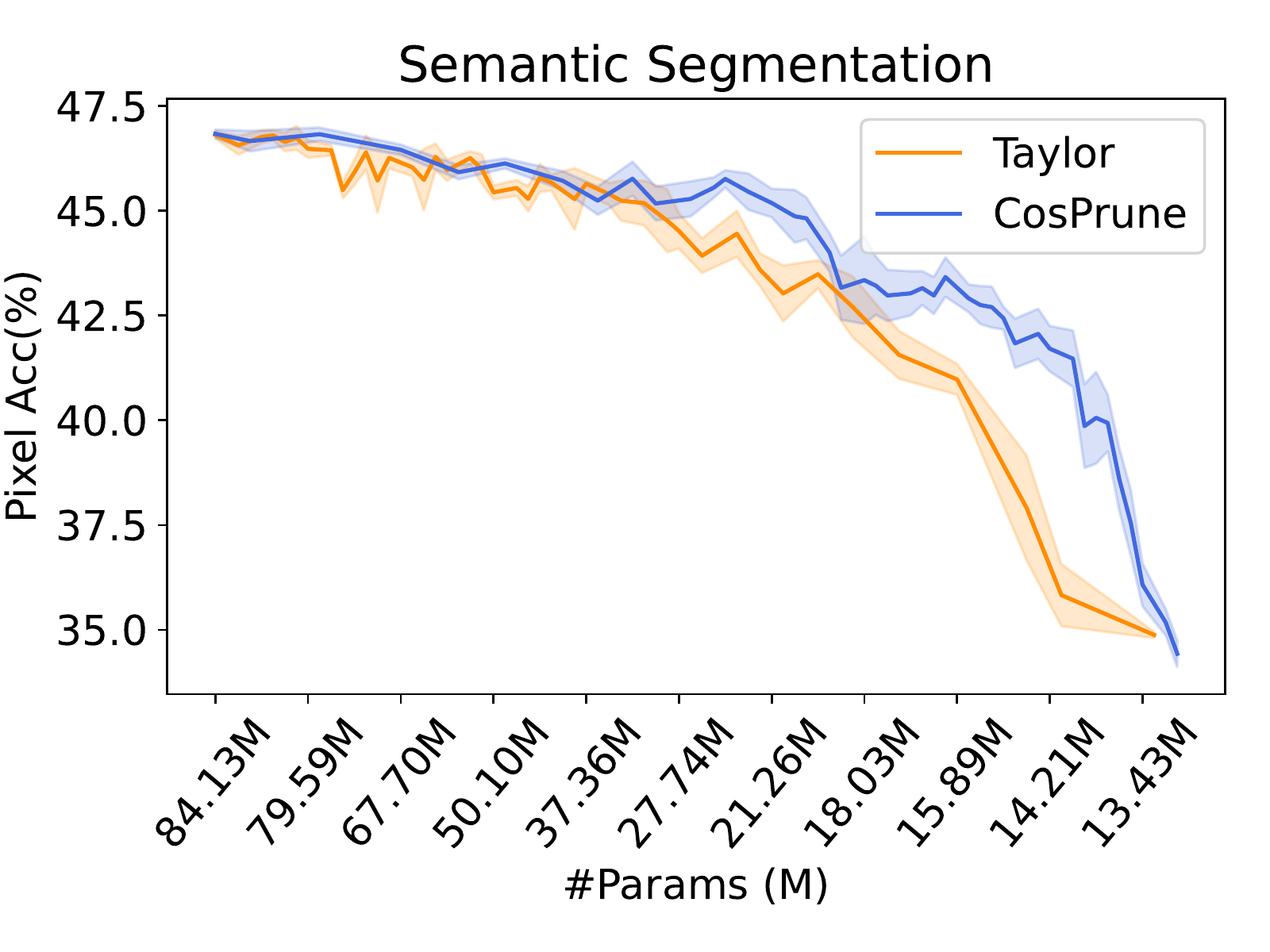}
    \caption{Pixel Accuracy}
    \label{fig:short-a-1}
  \end{subfigure}\hfill
  \begin{subfigure}{0.33\textwidth}
    \includegraphics[scale=0.35]{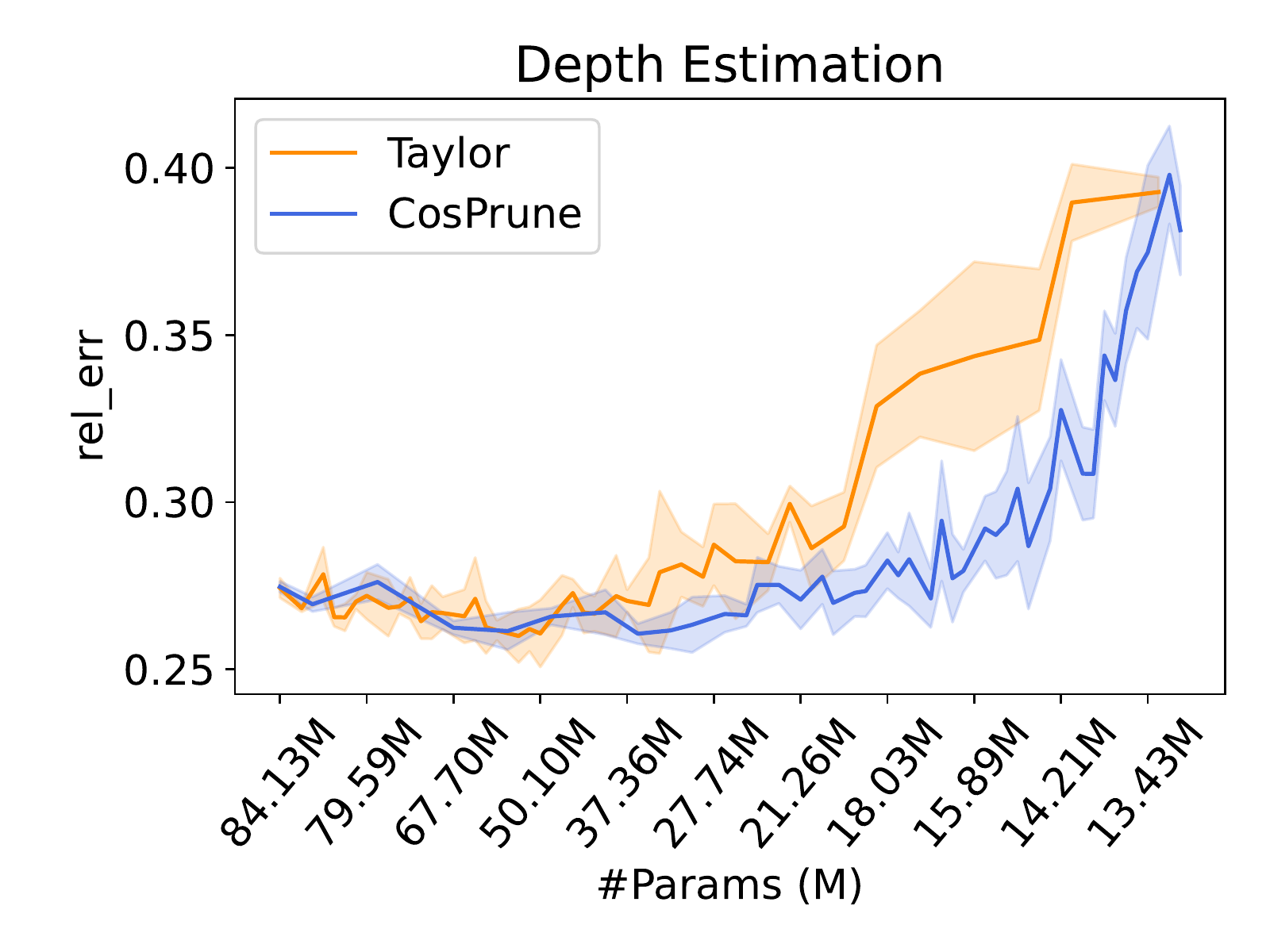}
    \caption{Relative Error}
    \label{fig:short-b-1}
  \end{subfigure}\hfill
  \begin{subfigure}{0.33\textwidth}
    \includegraphics[scale=0.35]{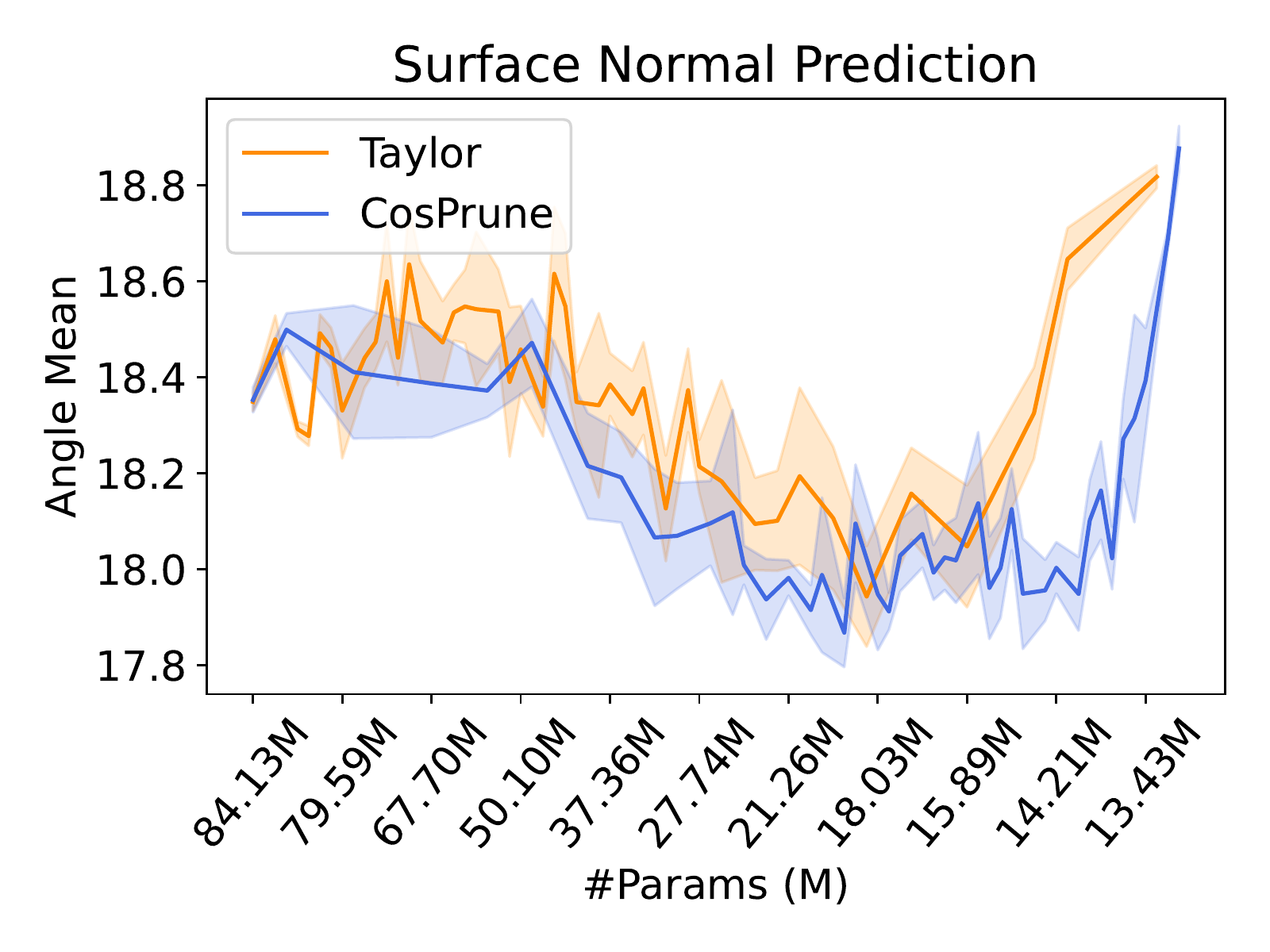}
    \caption{Angle Mean}
    \label{fig:short-c-1}
  \end{subfigure}
  \caption{Iterative pruning performance trends for different tasks for the Taylor and CosPrune methods.}
  \label{fig:random-init-iter}
\end{figure*}

\textbf{Model Pruning}: After the model is trained on the NYUv2 dataset, we apply iterative pruning with different filter pruning criteria (CosPrune and Taylor pruning) to get pruned MTL models. We initialize the model with the NYUv2 trained weights and set the initial learning rate as $10^{-5}$ with cosine scheduling for $500$ epochs and start the pruning process. For every iteration (batch of inputs to the model), we get the individual task gradients by applying backward propagation on task losses. Then we calculate the importance score for every filter in the model. The importance scores are accumulated for 10 epochs after which 100 filters that have the lowest accumulated scores, are pruned. This completes a single pruning iteration and after it, the accumulated filter importance scores are reset to $0$. The learning rate and scheduler are also rewinded to the initial settings and the pruning process continues. Note that the parameters of the model keep on updating using the total loss function after every iteration which marks the fine-tuning phase of the model. 

\subsection{Results: Multi-Task Structure Pruning}

\textbf{Iterative Pruning Results}: We applied the iterative pruning strategy with both the Taylor Pruning criterion and the CosPrune importance criterion separately on the model initialized with the trained NYUv2 weights and ran $6$ sets of experiments  to accommodate for the variance in each method.  We present a subset of evaluation metrics in Figure \ref{fig:random-init-iter}, where we show the best values for different task metrics ($y$-axis) against the number of parameters ($x$-axis) for both methods. We have shown plots for Pixel Accuracy (Semantic Segmentation), relative error (Depth Estimation), and Angle Mean (Normal prediction). The plots for the remaining metrics are reported in the supplementary material. 

From the plots in Figure \ref{fig:random-init-iter}, we can see that the CosPrune method is consistently better than the Taylor pruning method across all the tasks, i.e., it has a relatively lower Pixel Accuracy drop in the case of Semantic Segmentation, and lower error increment in case of Depth Estimation and Normal prediction tasks. We also report numerical values in Table \ref{tab:random-iter-init-all} where we compare both the methods for the pruned models with a similar number of parameters. We can see that there is the highest performance gap at $14.3M$ model parameters which is $83\%$ parameter reduction, and Pixel Accuracy is $\mathbf{16.39\%}$ better, Depth Estimation is $\mathbf{15.93\%}$ better and Normal prediction is $\mathbf{3.45\%}$ better than the Taylor pruning method. However, when with extreme pruning at $13.4M$ parameters, the gap is significantly lesser but nonetheless still better. For $59.9M$ parameters, the Pixel Accuracy slightly drops for CosPrune but all the metrics are very close as can also be seen from the plots in Figure \ref{fig:random-init-iter}. 

\begin{table*}[htb]
  \centering
  \begin{tabular}
{P{0.04\linewidth}P{0.04\linewidth}|P{0.07\linewidth}P{0.07\linewidth}|P{0.07\linewidth}P{0.07\linewidth}|P{0.07\linewidth}P{0.07\linewidth}|P{0.07\linewidth}P{0.07\linewidth}}
    \toprule
    \multicolumn{2}{P{0.08\linewidth}|}{\multirow{2}{*}{$\#$Param}} & \multicolumn{2}{|P{0.14\linewidth}|}{\multirow{2}{*}{$lr$ setting}} & \multicolumn{2}{|p{0.2\linewidth}|}{\centering Semantic Segmentation \\Pixel Accuracy $(\%)$ $\uparrow$} & \multicolumn{2}{|p{0.16\linewidth}|}{\centering Depth Estimation \\ Relative Error $\downarrow$} & \multicolumn{2}{|p{0.16\linewidth}}{\centering Surface Normal \\ Angle Mean $\downarrow$} \\
    \midrule
    $\# (M)$ & $\%_R$ & Taylor & CosPrune & Taylor & CosPrune & Taylor &  CosPrune & Taylor & CosPrune  \\
    \midrule
    $60.3$ & $28.3$ & $0.0005$ & $0.0001$ & $\textbf{47.82}$ & $46.52$ & $0.2688$ & $\textbf{0.2525}$ & $\textbf{17.97}$ & $18.20$\\
    $37.1$ & $55.8$ & $0.001\phantom{1}$ & $0.0001$ & $\textbf{46.51}$ & $46.18$ & $0.2761$ & $\textbf{0.2614}$ & $\textbf{17.91}$ & $18.21$\\
    $29.7$ & $64.6$ &  $0.001\phantom{1}$ & $0.0001$ & $\textbf{46.44}$ & $45.34$ & $0.2835$ & $\textbf{0.2761}$ & $\textbf{17.85}$ & $18.09$\\
    $25.9$ & $69.2$ &  $0.001\phantom{1}$ & $0.0005$ & $46.86$ & $\mathbf{47.84}$ & $0.2853$ & $\mathbf{0.2604}$ & $\mathbf{17.78}$ & $17.97$\\
    $19.1$ & $77.3$ &  $0.001\phantom{1}$ & $0.0001$ & $\mathbf{45.54}$ & $43.42$ & $0.2874$ & $\mathbf{0.2570}$ & $\mathbf{17.86}$ & $17.92$\\
    $15.3$ & $81.8$ &  $0.001\phantom{1}$ & $0.0001$ & $42.86$ & $\mathbf{44.67}$ & $0.3362$ & $\mathbf{0.2957}$ & $18.08$ & $\mathbf{17.99}$\\
    $14.4$ & $82.9$ &  $0.001\phantom{1}$ & $0.001\phantom{1}$ & $41.82$ & $\mathbf{44.40}$ & $0.3381$ & $\mathbf{0.3109}$ & $18.10$ & $\mathbf{18.06}$\\
    \bottomrule
  \end{tabular}
  \caption{Results of the pruned models trained from scratch across different tasks. $\# (M)$: \emph{Number of model parameters in millions}. $\%_R$: \emph{$\%$ parameter reduction with respect to full model}. $lr$ setting: \emph{Optimal learning rate for the respective model}.  $\uparrow$: \emph{Higher the better} $\downarrow$: \emph{Lower the better}. }
  \label{tab:pruned-models}
\end{table*}

\begin{figure*}[htb]
  \centering
  \begin{subfigure}{0.33\textwidth}
    \includegraphics[scale=0.35]{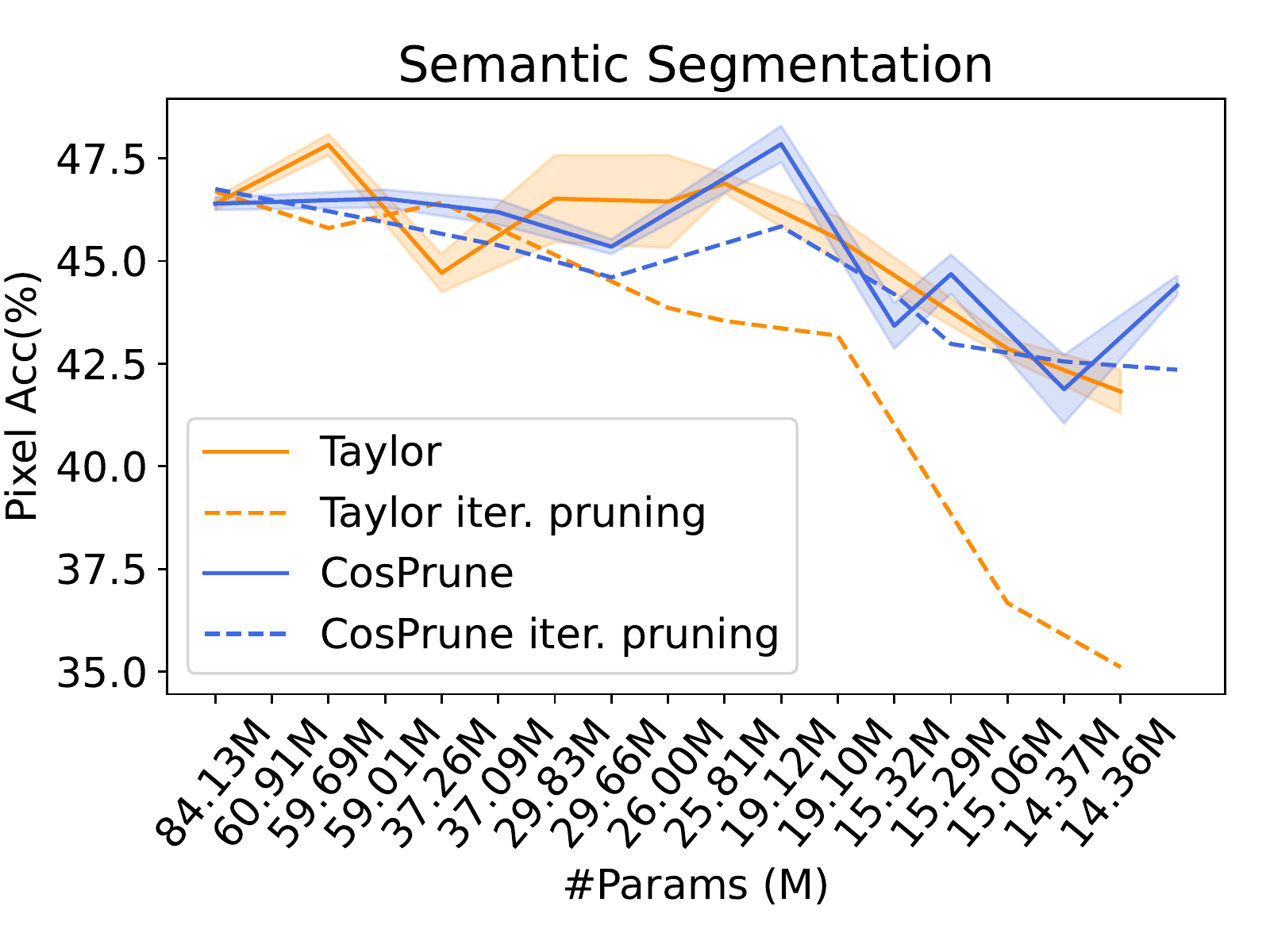}
    \caption{Pixel Accuracy}
    \label{fig:short-a}
  \end{subfigure}\hfill
  \begin{subfigure}{0.33\textwidth}
    \includegraphics[scale=0.35]{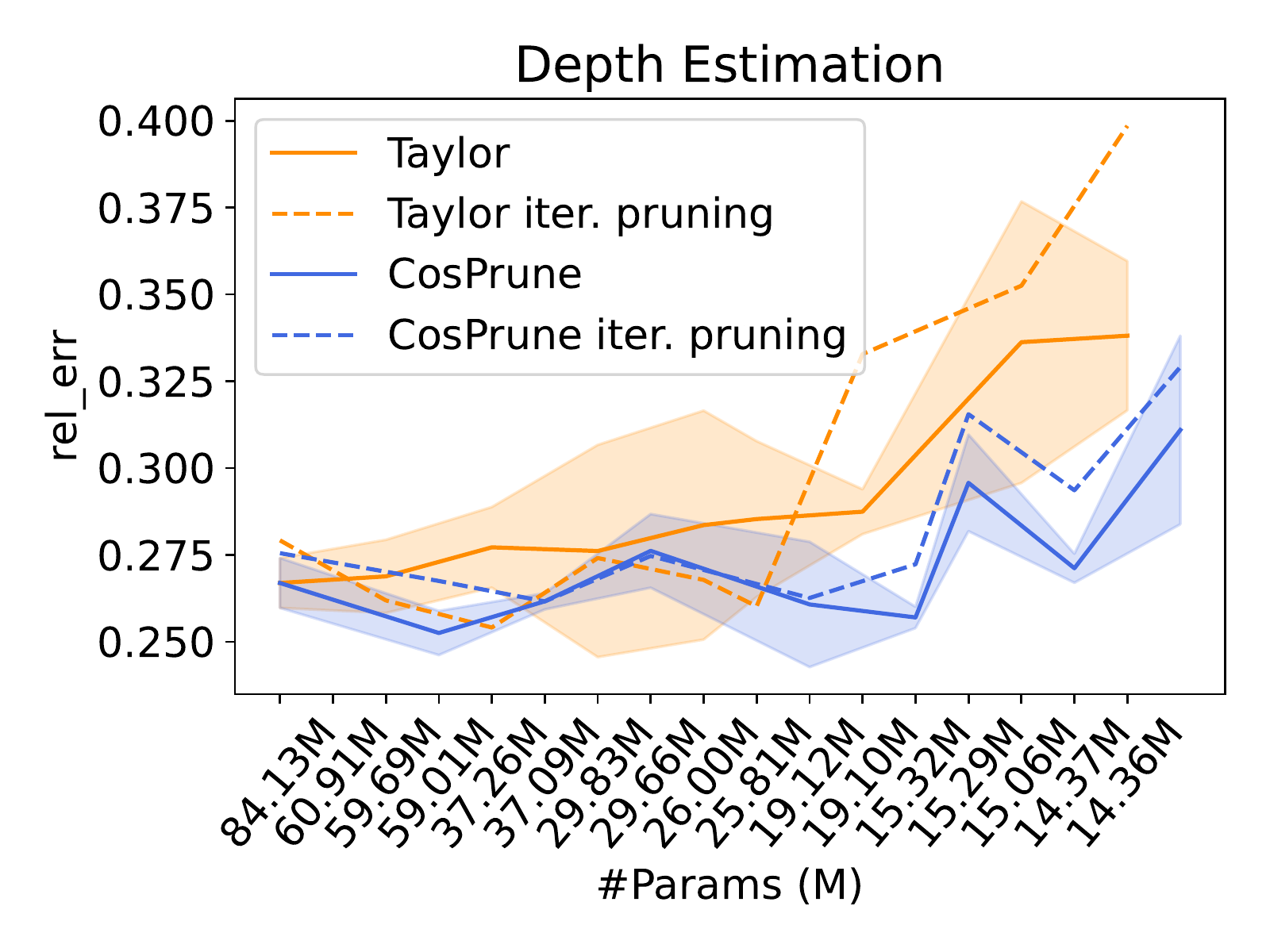}
    \caption{Relative Error}
    \label{fig:short-b}
  \end{subfigure}\hfill
  \begin{subfigure}{0.33\textwidth}
    \includegraphics[scale=0.35]{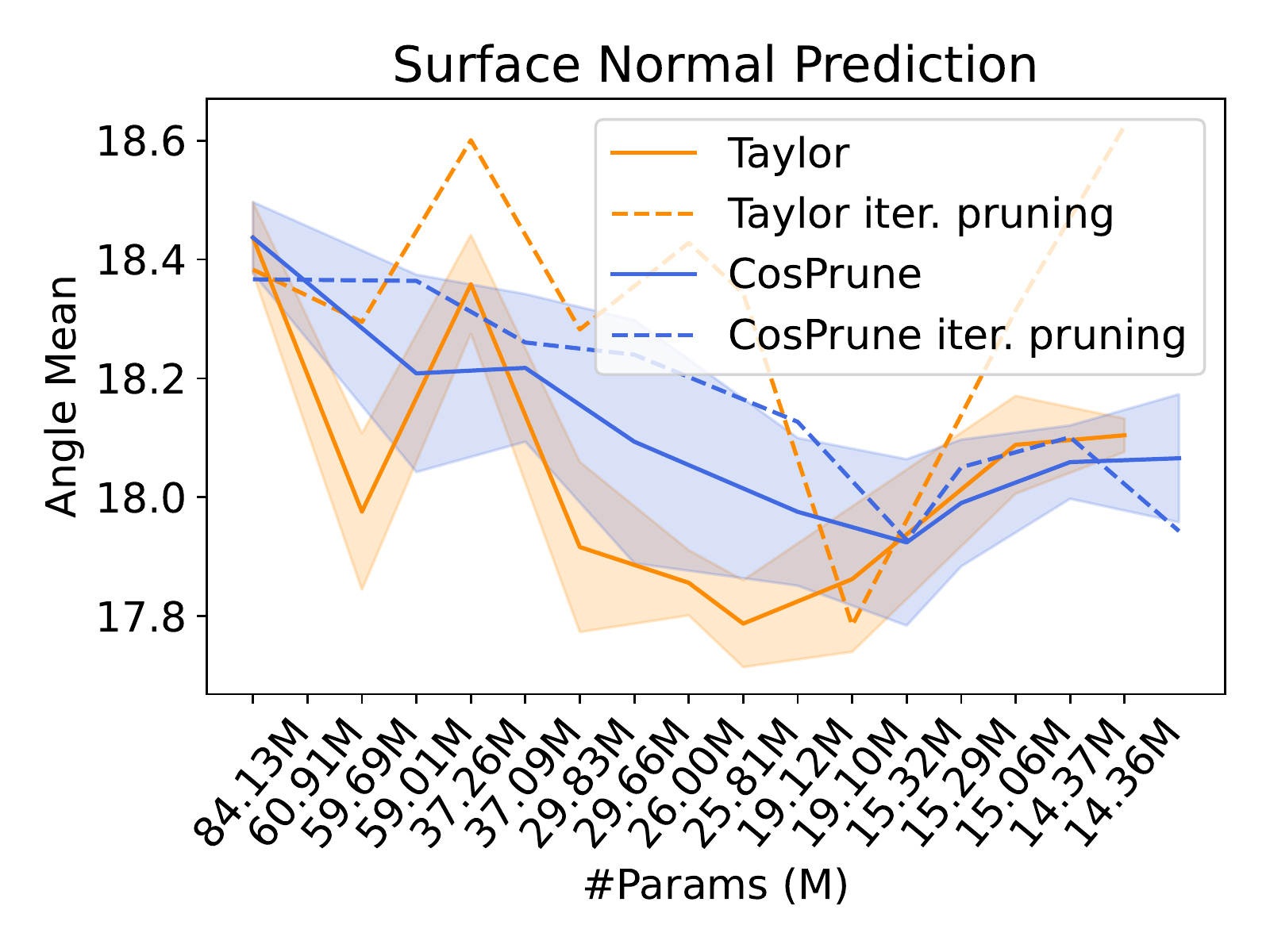}
    \caption{Angle Mean}
    \label{fig:short-c}
  \end{subfigure}
  \caption{Results of the pruned models, trained individually from scratch, across all the tasks for various model parameters. Solid plot lines correspond to primary results. Dashed plots correspond to the iterative pruning trends from Figure \ref{fig:random-init-iter}.}
  \label{fig:pruned-models}
\end{figure*}

\textbf{Re-training the Pruned models}: So far we have observed that the CosPrune can efficiently prune the MTL model to achieve high parameter reduction within a few iterations while maintaining its performance over the baseline method. Both pruning criteria lead to pruned models with different architectural designs in terms of the number of filters in each layer of the backbone. Since there is a difference in the performance, we want to figure out the role of these different layer configurations in the model's performance for all the tasks. To do this, we take the pruned models at various iterations, randomly initialize the weights and train them to converge on the NYUv2 dataset. Initially, we were using the same hyperparameters for training all the models but saw that some models were severely affected. For example, a model with $15M$ parameters, obtained from CosPrune could achieve only $25.5\%$ Pixel Accuracy even after training for $500$ epochs. But when the learning rate of $0.0001$ was used it reached over $42\%$, and the performance of the other two tasks was also improved. In another case, a model obtained from Taylor Pruning with $14M$ parameters resulted in $38\%$ Pixel Accuracy with $0.0001$ learning rate, whereas the same model reached over $44\%$ with $0.001$ learning rate. Moreover, some pruned models gave their best results for $0.0005$ learning rate. 

\begin{table*}[htb]
  \centering
  \small 
  \tabcolsep=0.05cm
  \begin{tabular}{l|P{0.12\linewidth}|P{0.115\linewidth}|P{0.115\linewidth}|P{0.115\linewidth}|P{0.115\linewidth}|P{0.115\linewidth}}
    \toprule
    \multirow{2}{*}{\textbf{Validation Metric}}  & \multicolumn{2}{|P{0.23\linewidth}}{Semantic Segmentation} & \multicolumn{2}{|P{0.23\linewidth}}{Depth Estimation} & \multicolumn{2}{|P{0.23\linewidth}}{Surface Normal} \\
    \cmidrule{2-7} 
    & Pix.Acc.$(\%)$ $\uparrow$ & mIoU $\uparrow$ & Abs.Err. $\downarrow$ & Rel.Err $\downarrow$ & Mean $\downarrow$ & Median $\downarrow$\\
    \midrule
    
    Total Validation Loss & $42.81$ & $0.1526$ & $0.6627$ & $0.2764$ & $\mathbf{18.08}$ & $16.15$\\ 
    Pixel Accuracy & $\mathbf{46.56}$ & $\mathbf{0.1654}$ & $\mathbf{0.6239}$ & $\mathbf{0.2527}$ & $18.51$ & $\mathbf{15.55}$  \\
    \bottomrule
  \end{tabular}
  \caption{Results at the best training epoch when different validation metrics are used.}
  \label{tab:val-metric}
\end{table*}

Therefore, to get the best possible performance from every pruned model, we trained them on $3$ sets of learning rates which are $0.001$, $0.0005$, and $0.0001$. We ran all the experiments for $500$ epochs with cosine scheduling and used Adam \cite{kingma2014adam} optimizer for training. We summarize results for a subset of models in Table \ref{tab:pruned-models}. Each row shows the performance comparison between pruned models obtained from Taylor pruning and CosPrune which have the same number of parameters. We have also given the desired learning rate for each model. It seems that Taylor pruned models favor $0.001$ learning rate with a few favoring $0.0001$, and $0.0005$. On the other hand, CosPruned models favor smaller learning rates of $0.0001$, and $0.0005$. The corresponding plots for the 3 tasks are shown in Figure \ref{fig:pruned-models}.

From the results in Table \ref{tab:pruned-models}, we can see that the models obtained from different pruning methods lead to very similar performance across all the tasks when the number of parameters is roughly equal. Figure \ref{fig:pruned-models}, also shows this similar trend where the winning method keeps on fluctuating with changing model parameters. In some cases, Taylor pruned models perform better than CosPruned models and in other cases, it is vice-versa but the results are still very close for both methods across all tasks. We have also compared the results of the individually pruned models with 
 the results from the corresponding iterative pruning iterations in Figure \ref{fig:pruned-models}. It can be seen that at very low parameters, the iterative pruning results become worse than the results of the models that were trained solely from scratch except in the case of normal estimation where the Angle Mean is slightly better in the case of the iterative CosPrune method.

\textbf{Analyzing the Results}: We observed that if we consider the pruned models obtained from two different pruning methods, randomly initialize them and re-train them with their optimal learning rate, they give similar results for the same number of parameters. At some parameter levels, the pruned model obtained from Taylor pruning performed better but at some other parameter levels, the pruned model obtained from CosPrune gave better results. But there is only a slight difference in the results which could be induced by the randomness of the training process. 
By looking at these results, we can conclude that different architecture configurations with a similar number of parameters do not play a major role in the final performance across tasks as long as the models are trained with their optimal training settings. 

Furthermore, we saw that the structured iterative pruning was biased towards one pruning criterion over the other as Figure \ref{fig:random-init-iter} shows that the CosPrune models outperformed Taylor pruned models at the same parameter level. 
But when the pruned models were re-trained from scratch, their final performances became similar and even closer to the unpruned model as seen in Figure \ref{fig:pruned-models}.
But for both methods, there was a steep drop in the performance at extreme pruning ratios, which was again mitigated by random initialization and re-training that led to better performance even at those extreme pruning ratios. 
This might be happening because, in the case of iterative structured pruning, the inherited trained weights from previous pruning iterations may act as bad initialization for the pruned models at successive iterations \cite{liu2018rethinking}. Even if those weights were optimal for the earlier unpruned model versions, they might not be good for the current version because of changes in the architecture after pruning and it can be difficult for the optimizer to find another good local minimum. 

\subsection{Additional Observation for NYUv2 Dataset}

\textbf{Best Validation Metric}: In the case of multi-task learning, total validation loss is traditionally considered a good validation metric that implies that the model is performing well on all the tasks collectively. However, in our experiments on the NYUv2 dataset, we observed that most of the time the least total validation loss failed to achieve the best model performance. In many instances, the Pixel Accuracy was low at the epoch with the best validation loss but it reached higher at some other epochs but with a slightly higher loss. However, the results for Depth Estimation and Surface Normal Prediction were similar at these different epochs. To verify this further, we tracked the best Pixel Accuracy in a training run and also tracked the best total validation loss in a separate training run but done with the same settings of hyperparameters. We trained the model 3 times to accommodate for the variance and show these results in Table \ref{tab:val-metric}. We can see that all the metric values except the Angle Mean are better when the model at the highest Pixel Accuracy is considered as compared to the model at least validation loss. This is also quite reasonable because the Semantic Segmentation task is the most difficult task and the scale of its loss function is higher than the other two tasks. Therefore, for getting the best model, we considered the best Pixel Accuracy score instead of the total validation loss.

\section{Conclusions}

In this work, we used two different structured pruning methods to compress deep multi-task neural networks. We are first to comprehensively analyze their effectiveness in pruning such networks. Specifically, we showed that different architectural configurations of the pruned models can give similar results regardless of the pruning method used if the number of parameters is the same. We also showed that iterative structured pruning may not be the best strategy to compress deep multi-task models. They might favor one pruning strategy over another. But regardless of the pruning strategy used, the performance of the pruned models can take a steep drop after certain iterations. However, we also showed that after random initialization and re-training those models with their respective optimal learning rates, they can give much higher performance across all the tasks. 



{\small
\bibliographystyle{ieee_fullname}
\bibliography{egbib}

\begin{thebibliography}{10}\itemsep=-1pt

\bibitem{abdollahzadeh2021revisit-knowledge}
Milad Abdollahzadeh, Touba Malekzadeh, and Ngai-Man~Man Cheung.
\newblock Revisit multimodal meta-learning through the lens of multi-task
  learning.
\newblock {\em Advances in Neural Information Processing Systems},
  34:14632--14644, 2021.

\bibitem{caruana1993multitask-hard-1}
R Caruana.
\newblock Multitask learning: A knowledge-based source of inductive bias1.
\newblock In {\em Proceedings of the Tenth International Conference on Machine
  Learning}, pages 41--48. Citeseer, 1993.

\bibitem{aspp}
Liang-Chieh Chen, George Papandreou, Iasonas Kokkinos, Kevin Murphy, and Alan~L
  Yuille.
\newblock Semantic image segmentation with deep convolutional nets and fully
  connected crfs.
\newblock {\em arXiv preprint arXiv:1412.7062}, 2014.

\bibitem{cheng2021multi-filter-index-share}
Hanjing Cheng, Zidong Wang, Lifeng Ma, Xiaohui Liu, and Zhihui Wei.
\newblock Multi-task pruning via filter index sharing: A many-objective
  optimization approach.
\newblock {\em Cognitive Computation}, 13:1070--1084, 2021.

\bibitem{cheng2021multi-merge-1}
Hanjing Cheng, Zidong Wang, Lifeng Ma, Xiaohui Liu, and Zhihui Wei.
\newblock Multi-task pruning via filter index sharing: A many-objective
  optimization approach.
\newblock {\em Cognitive Computation}, 13:1070--1084, 2021.

\bibitem{opt-1-desideri2012multiple}
Jean-Antoine D{\'e}sid{\'e}ri.
\newblock Multiple-gradient descent algorithm (mgda) for multiobjective
  optimization.
\newblock {\em Comptes Rendus Mathematique}, 350(5-6):313--318, 2012.

\bibitem{ding2021resrep-learn-1}
Xiaohan Ding, Tianxiang Hao, Jianchao Tan, Ji Liu, Jungong Han, Yuchen Guo, and
  Guiguang Ding.
\newblock Resrep: Lossless cnn pruning via decoupling remembering and
  forgetting.
\newblock In {\em Proceedings of the IEEE/CVF International Conference on
  Computer Vision}, pages 4510--4520, 2021.

\bibitem{nyudepthnormal}
David Eigen and Rob Fergus.
\newblock Predicting depth, surface normals and semantic labels with a common
  multi-scale convolutional architecture.
\newblock In {\em Proceedings of the IEEE international conference on computer
  vision}, pages 2650--2658, 2015.

\bibitem{depth-threshold}
David Eigen, Christian Puhrsch, and Rob Fergus.
\newblock Depth map prediction from a single image using a multi-scale deep
  network.
\newblock {\em Advances in neural information processing systems}, 27, 2014.

\bibitem{rigl-evci2020rigging}
Utku Evci, Trevor Gale, Jacob Menick, Pablo~Samuel Castro, and Erich Elsen.
\newblock Rigging the lottery: Making all tickets winners.
\newblock In {\em International Conference on Machine Learning}, pages
  2943--2952. PMLR, 2020.

\bibitem{lth-1}
Jonathan Frankle and Michael Carbin.
\newblock The lottery ticket hypothesis: Finding sparse, trainable neural
  networks.
\newblock {\em arXiv preprint arXiv:1803.03635}, 2018.

\bibitem{wt-rewind}
Jonathan Frankle, Gintare~Karolina Dziugaite, Daniel~M Roy, and Michael Carbin.
\newblock The lottery ticket hypothesis at scale.
\newblock {\em arXiv preprint arXiv:1903.01611}, 2020.

\bibitem{ghiasi2021multi-cv3}
Golnaz Ghiasi, Barret Zoph, Ekin~D Cubuk, Quoc~V Le, and Tsung-Yi Lin.
\newblock Multi-task self-training for learning general representations.
\newblock In {\em Proceedings of the IEEE/CVF International Conference on
  Computer Vision}, pages 8856--8865, 2021.

\bibitem{guo2018dynamic-cv2}
Michelle Guo, Albert Haque, De-An Huang, Serena Yeung, and Li Fei-Fei.
\newblock Dynamic task prioritization for multitask learning.
\newblock In {\em Proceedings of the European conference on computer vision
  (ECCV)}, pages 270--287, 2018.

\bibitem{weights-and-connections}
Song Han, Jeff Pool, John Tran, and William Dally.
\newblock Learning both weights and connections for efficient neural network.
\newblock {\em Advances in neural information processing systems}, 28, 2015.

\bibitem{brain-surgeon-second-derivative}
Babak Hassibi and David Stork.
\newblock Second order derivatives for network pruning: Optimal brain surgeon.
\newblock {\em Advances in neural information processing systems}, 5, 1992.

\bibitem{he2021pruning-pam}
Xiaoxi He, Dawei Gao, Zimu Zhou, Yongxin Tong, and Lothar Thiele.
\newblock Pruning-aware merging for efficient multitask inference.
\newblock In {\em Proceedings of the 27th ACM SIGKDD Conference on Knowledge
  Discovery \& Data Mining}, pages 585--595, 2021.

\bibitem{he2018multi-merge-2}
Xiaoxi He, Zimu Zhou, and Lothar Thiele.
\newblock Multi-task zipping via layer-wise neuron sharing.
\newblock {\em Advances in Neural Information Processing Systems}, 31, 2018.

\bibitem{hou2022chex-learn-2}
Zejiang Hou, Minghai Qin, Fei Sun, Xiaolong Ma, Kun Yuan, Yi Xu, Yen-Kuang
  Chen, Rong Jin, Yuan Xie, and Sun-Yuan Kung.
\newblock Chex: channel exploration for cnn model compression.
\newblock In {\em Proceedings of the IEEE/CVF Conference on Computer Vision and
  Pattern Recognition}, pages 12287--12298, 2022.

\bibitem{hu2021unit-cv1}
Ronghang Hu and Amanpreet Singh.
\newblock Unit: Multimodal multitask learning with a unified transformer.
\newblock In {\em Proceedings of the IEEE/CVF International Conference on
  Computer Vision}, pages 1439--1449, 2021.

\bibitem{kendall2018multi-c2}
Alex Kendall, Yarin Gal, and Roberto Cipolla.
\newblock Multi-task learning using uncertainty to weigh losses for scene
  geometry and semantics.
\newblock In {\em Proceedings of the IEEE conference on computer vision and
  pattern recognition}, pages 7482--7491, 2018.

\bibitem{kim2021mila-latency}
Donghyun Kim, Tian Lan, Chuhang Zou, Ning Xu, Bryan~A Plummer, Stan Sclaroff,
  Jayan Eledath, and Gerard Medioni.
\newblock Mila: Multi-task learning from videos via efficient inter-frame
  attention.
\newblock In {\em Proceedings of the IEEE/CVF International Conference on
  Computer Vision}, pages 2219--2229, 2021.

\bibitem{kingma2014adam}
Diederik~P Kingma and Jimmy Ba.
\newblock Adam: A method for stochastic optimization.
\newblock {\em arXiv preprint arXiv:1412.6980}, 2014.

\bibitem{optimal=brain-damage}
Yann LeCun, John Denker, and Sara Solla.
\newblock Optimal brain damage.
\newblock {\em Advances in neural information processing systems}, 2, 1989.

\bibitem{lee2018snip}
Namhoon Lee, Thalaiyasingam Ajanthan, and Philip~HS Torr.
\newblock Snip: Single-shot network pruning based on connection sensitivity.
\newblock {\em arXiv preprint arXiv:1810.02340}, 2018.

\bibitem{pruning-filters}
Hao Li, Asim Kadav, Igor Durdanovic, Hanan Samet, and Hans~Peter Graf.
\newblock Pruning filters for efficient convnets.
\newblock {\em arXiv preprint arXiv:1608.08710}, 2016.

\bibitem{random-li2022revisiting}
Yawei Li, Kamil Adamczewski, Wen Li, Shuhang Gu, Radu Timofte, and Luc
  Van~Gool.
\newblock Revisiting random channel pruning for neural network compression.
\newblock In {\em Proceedings of the IEEE/CVF Conference on Computer Vision and
  Pattern Recognition}, pages 191--201, 2022.

\bibitem{opt-2-liu2021conflict}
Bo Liu, Xingchao Liu, Xiaojie Jin, Peter Stone, and Qiang Liu.
\newblock Conflict-averse gradient descent for multi-task learning.
\newblock {\em Advances in Neural Information Processing Systems},
  34:18878--18890, 2021.

\bibitem{random-liu2022unreasonable}
Shiwei Liu, Tianlong Chen, Xiaohan Chen, Li Shen, Decebal~Constantin Mocanu,
  Zhangyang Wang, and Mykola Pechenizkiy.
\newblock The unreasonable effectiveness of random pruning: Return of the most
  naive baseline for sparse training.
\newblock {\em arXiv preprint arXiv:2202.02643}, 2022.

\bibitem{liu2018rethinking}
Zhuang Liu, Mingjie Sun, Tinghui Zhou, Gao Huang, and Trevor Darrell.
\newblock Rethinking the value of network pruning.
\newblock {\em arXiv preprint arXiv:1810.05270}, 2018.

\bibitem{cosineschedule}
Ilya Loshchilov and Frank Hutter.
\newblock Sgdr: Stochastic gradient descent with warm restarts.
\newblock {\em arXiv preprint arXiv:1608.03983}, 2016.

\bibitem{adamw}
Ilya Loshchilov and Frank Hutter.
\newblock Decoupled weight decay regularization.
\newblock {\em arXiv preprint arXiv:1711.05101}, 2017.

\bibitem{bayesian-compression}
Christos Louizos, Karen Ullrich, and Max Welling.
\newblock Bayesian compression for deep learning.
\newblock {\em Advances in neural information processing systems}, 30, 2017.

\bibitem{lu2017fully-c1}
Yongxi Lu, Abhishek Kumar, Shuangfei Zhai, Yu Cheng, Tara Javidi, and Rogerio
  Feris.
\newblock Fully-adaptive feature sharing in multi-task networks with
  applications in person attribute classification.
\newblock In {\em Proceedings of the IEEE conference on computer vision and
  pattern recognition}, pages 5334--5343, 2017.

\bibitem{misra2016cross-general}
Ishan Misra, Abhinav Shrivastava, Abhinav Gupta, and Martial Hebert.
\newblock Cross-stitch networks for multi-task learning.
\newblock In {\em Proceedings of the IEEE conference on computer vision and
  pattern recognition}, pages 3994--4003, 2016.

\bibitem{taylor-pruning}
Pavlo Molchanov, Arun Mallya, Stephen Tyree, Iuri Frosio, and Jan Kautz.
\newblock Importance estimation for neural network pruning.
\newblock In {\em Proceedings of the IEEE/CVF conference on computer vision and
  pattern recognition}, pages 11264--11272, 2019.

\bibitem{sd-mtl-1-phillips2021deep}
John Phillips, Julieta Martinez, Ioan~Andrei B{\^a}rsan, Sergio Casas, Abbas
  Sadat, and Raquel Urtasun.
\newblock Deep multi-task learning for joint localization, perception, and
  prediction.
\newblock In {\em Proceedings of the IEEE/CVF Conference on Computer Vision and
  Pattern Recognition}, pages 4679--4689, 2021.

\bibitem{ruder122017learning-c3}
Sebastian Ruder12, Joachim Bingel, Isabelle Augenstein, and Anders S{\o}gaard.
\newblock Learning what to share between loosely related tasks.
\newblock {\em arXiv preprint arXiv:1705.08142}, 2017.

\bibitem{sener2018multi-hard-2}
Ozan Sener and Vladlen Koltun.
\newblock Multi-task learning as multi-objective optimization.
\newblock {\em Advances in neural information processing systems}, 31, 2018.

\bibitem{nyusegment}
Nathan Silberman, Derek Hoiem, Pushmeet Kohli, and Rob Fergus.
\newblock Indoor segmentation and support inference from rgbd images.
\newblock {\em ECCV (5)}, 7576:746--760, 2012.

\bibitem{vgg}
Karen Simonyan and Andrew Zisserman.
\newblock Very deep convolutional networks for large-scale image recognition.
\newblock {\em arXiv preprint arXiv:1409.1556}, 2014.

\bibitem{rl-mtl-2-sodhani2021multi}
Shagun Sodhani, Amy Zhang, and Joelle Pineau.
\newblock Multi-task reinforcement learning with context-based representations.
\newblock In {\em International Conference on Machine Learning}, pages
  9767--9779. PMLR, 2021.

\bibitem{sun2022disparse}
Xinglong Sun, Ali Hassani, Zhangyang Wang, Gao Huang, and Humphrey Shi.
\newblock Disparse: Disentangled sparsification for multitask model
  compression.
\newblock In {\em Proceedings of the IEEE/CVF Conference on Computer Vision and
  Pattern Recognition}, pages 12382--12392, 2022.

\bibitem{sun2020adashare}
Ximeng Sun, Rameswar Panda, Rogerio Feris, and Kate Saenko.
\newblock Adashare: Learning what to share for efficient deep multi-task
  learning.
\newblock {\em Advances in Neural Information Processing Systems},
  33:8728--8740, 2020.

\bibitem{rl-mtl-1-teh2017distral}
Yee Teh, Victor Bapst, Wojciech~M Czarnecki, John Quan, James Kirkpatrick, Raia
  Hadsell, Nicolas Heess, and Razvan Pascanu.
\newblock Distral: Robust multitask reinforcement learning.
\newblock {\em Advances in neural information processing systems}, 30, 2017.

\bibitem{williams2008multi-robo-1}
Christopher Williams, Stefan Klanke, Sethu Vijayakumar, and Kian Chai.
\newblock Multi-task gaussian process learning of robot inverse dynamics.
\newblock {\em Advances in neural information processing systems}, 21, 2008.

\bibitem{wilson2007multi-rl-3}
Aaron Wilson, Alan Fern, Soumya Ray, and Prasad Tadepalli.
\newblock Multi-task reinforcement learning: a hierarchical bayesian approach.
\newblock In {\em Proceedings of the 24th international conference on Machine
  learning}, pages 1015--1022, 2007.

\bibitem{yang2020multi-rl-4}
Ruihan Yang, Huazhe Xu, Yi Wu, and Xiaolong Wang.
\newblock Multi-task reinforcement learning with soft modularization.
\newblock {\em Advances in Neural Information Processing Systems},
  33:4767--4777, 2020.

\bibitem{pcgrad-yu2020gradient}
Tianhe Yu, Saurabh Kumar, Abhishek Gupta, Sergey Levine, Karol Hausman, and
  Chelsea Finn.
\newblock Gradient surgery for multi-task learning.
\newblock {\em Advances in Neural Information Processing Systems},
  33:5824--5836, 2020.

\bibitem{yu2020meta-robo-2}
Tianhe Yu, Deirdre Quillen, Zhanpeng He, Ryan Julian, Karol Hausman, Chelsea
  Finn, and Sergey Levine.
\newblock Meta-world: A benchmark and evaluation for multi-task and meta
  reinforcement learning.
\newblock In {\em Conference on robot learning}, pages 1094--1100. PMLR, 2020.

\bibitem{sparse-hardware-zhu2019sparse}
Maohua Zhu, Tao Zhang, Zhenyu Gu, and Yuan Xie.
\newblock Sparse tensor core: Algorithm and hardware co-design for vector-wise
  sparse neural networks on modern gpus.
\newblock In {\em Proceedings of the 52nd Annual IEEE/ACM International
  Symposium on Microarchitecture}, pages 359--371, 2019.

\end{thebibliography}
}

\end{document}